\documentclass{article}
\usepackage{spconf,graphicx}
\usepackage{mathtools}
\usepackage{amssymb}
\usepackage{amsmath}
\usepackage{multirow}
\usepackage{algorithm}
\usepackage{algorithmic}
\usepackage{url}

\title{PROMPT LEARNING WITH KNOWLEDGE MEMORIZING PROTOTYPES FOR GENERALIZED FEW-SHOT INTENT DETECTION}
\name{Chaiyut Luoyiching$^{1}$, Yangning Li$^{1}$, Yinghui Li$^{1}$, Rongsheng Li$^{1}$, \textit{Hai-Tao Zheng}$^{1,2,*}$, \textit{Nannan Zhou}$^{3}$, \textit{Hanjing Su}$^{3}$\thanks{$^{*}$Corresponding author (E-mail: zheng.haitao@sz.tsinghua.edu.cn)}}
\address{
$^1$Shenzhen International Graduate School, Tsinghua University, Shenzhen, China;\\
$^2$Pengcheng Laboratory, Shenzhen, China;
$^3$Wechat Pay, Tencent, Shenzhen, China
}

\begin{document}
%\ninept
%
\maketitle
\begin{abstract}
Generalized Few-Shot Intent Detection (GFSID) is challenging and realistic because it needs to categorize both seen and novel intents simultaneously. Previous GFSID methods rely on the episodic learning paradigm, which makes it hard to extend to a generalized setup as they do not explicitly learn the classification of seen categories and the knowledge of seen intents. To address the dilemma, we propose to convert the GFSID task into the class incremental learning paradigm. Specifically, we propose a two-stage learning framework, which sequentially learns the knowledge of different intents in various periods via prompt learning.  And then we exploit prototypes for categorizing both seen and novel intents. Furthermore, to achieve the transfer knowledge of intents in different stages, for different scenarios we design two knowledge preservation methods which close to realistic applications. Extensive experiments and detailed analyses on two widely used datasets show that our framework based on the class incremental learning paradigm achieves promising performance.
\end{abstract}
\section{Introduction}
\label{sec:intro}

Intent Detection is a vital component in dialog systems, which categorizes user utterances into corresponding intent and then systems execute the corresponding instructions or other downstream tasks. In the real world, the systems have two major problems that need to be solved urgently: (1). Lacking a large amount of labeled data. Because annotating data is costly and the utterances of users are complex and diverse. (2). Systems require the addition of novel intents to accommodate more needs. So in this paper, we are interested in Generalized Few-Shot Intent Detection (GFSID), which is more challenging and realistic. In this setting, models need to classify both seen intents and novel intents simultaneously. 

For the problem of data scarcity \cite{DBLP:conf/emnlp/MaLSZHZLLLCZS22, DBLP:journals/corr/abs-2207-08087}, previous researchers have devoted many efforts in several different areas \cite{gao2019hybrid, sun2019hierarchical, ye2019multi, DBLP:journals/patterns/LiuLTLZ22, DBLP:conf/acl/LiZLLLSWLCZ22}. Some valuable backbone networks have emerged, such as prototype network \cite{snell2017prototypical}, matching network \cite{vinyals2016matching}, and relation network \cite{sung2018learning,DBLP:conf/sigir/LiLHYS022}. 

These above methods are hard to extend to a generalized setup directly where both seen and novel intents co-exist \cite{DBLP:journals/corr/abs-2211-04215}. In response to this situation, \cite{nguyen2020dynamic} propose a solution first in the field of intent detection. \cite{nguyen2020dynamic} considers that previous few-shot approaches rely on a static similarity measure and overly fine-grained matching components, which inhibit generalizing capability towards GFSID. So they propose a Semantic Matching and Aggregation Network which exploits additional knowledge by automatically extracting multiple semantic components. Besides, \cite{ijcai2022p617} thinks a few instances can not cover the diversity of user expressions, which leads to previous few-shot approaches falling shot in adapting to the generalized setup. To deal with this issue, \cite{ijcai2022p617} introduces a diversity feature enhanced prototypical network, which generates diversity features to enhance novel intents representation using an auxiliary set. However, both two methods ignore the learning of classification of seen categories and knowledge of seen intents.

To address the dilemma, we propose to convert the GFSID task into the class incremental learning paradigm and introduce a two-stage learning framework. Motivated by recent work around prompt learning, we design a predefined template for improving performance with the use of PLMs. Since prototypes have a good generalized capability \cite{snell2017prototypical} and are easy to transfer knowledge in vector space, we use prototypes for classification toward both seen and novel intents. During training, our framework first learns the knowledge of seen intents with a large amount of labeled data. And then in the second phase, the framework learns the knowledge of novel intents with a few labeled data. In the meantime, we use two methods to memorize the prototypes of seen intents and transfer old knowledge for the limitations of different application scenarios.

The main contributions of our work are as follows:
\begin{itemize}
\item We innovatively convert the GFSID task into the class incremental learning paradigm and propose a framework for learning knowledge of seen and novel intents. 
\item To close to realistic applications, we present two knowledge preservation methods for memorizing the prototypes of seen intents and transferring old knowledge for different scenarios.
\item We evaluate our framework in two publicly available intent detection datasets. Extensive experiments show that our framework significantly outperforms previous methods in most experimental settings.
\end{itemize}

\section{RELATED WORK}
\label{sec:format}

\textbf{Generalized Few-Shot Learning}. Few-shot learning (FSL) methods relied on a meta-learning paradigm, which trained models to recognize novel classes with a limited number of samples. Seminal methods for FSL were introduced\cite{snell2017prototypical, vinyals2016matching, sung2018learning}. Building upon these foundations, subsequent research has explored multi-level matching, aggregation approaches and adversarial adaptation networks \cite{gao2019hybrid, sun2019hierarchical, ye2019multi, han2021meta, DBLP:journals/corr/abs-2306-12245}. Generalized Few-Shot Learning(GFSL) represents a novel research direction that is still being explored. \cite{nguyen2020dynamic} leverage semantic knowledge to provide additional information, while \cite{ijcai2022p617} generate diverse features to enhance the representation of novel intents using an auxiliary set. \cite{gidaris2018dynamic} propose a system capable of learning novel categories with limited training data while retaining knowledge of previously learned base classes.

\noindent\textbf{Prompting PLM}. Inspired by GPT-3 \cite{brown2020language}, many researchers have employed ``Prompts'' or ``Templates'' to establish connections between pre-training tasks and downstream tasks\cite{schick2021exploiting, gao2021making, lester2021power, DBLP:journals/corr/abs-2307-09007}. Another crucial component in prompt learning is the "Verbalizer," which maps the model output to the corresponding label \cite{schick2021exploiting, gao2021making}.

\section{METHODOLOGY}
\label{sec:pagestyle}

% Within this section, we propose a prompt-based learning approach for intent detection and provide a detailed explanation of how to implement classification using prototypes. Furthermore, we present two techniques for knowledge preservation, enabling us to retain prototypes of previously encountered intents and preserve old knowledge for diverse applications. Transitioning to the third section, we introduce supervised contrastive learning as a means to augment the representation of user utterances. Lastly, we offer a comprehensive description of the entire framework and provide a detailed elaboration of the learning procedure.

\subsection{Prompt-based Learning for Intent Detection}
% Prompt learning is an approach that incorporates supplementary information, referred to as a "template," into the original sentence, transforming the downstream task into a cloze-style mask language modeling problem. The template comprises a few tokens that offer additional guidance to PLMs. Verbalizer serves as a key component in prompt learning, projecting the original labels onto a set of descriptive words.

% In this paper, we denote the input sentence as  $x=[w_1, ..., w_n]$, the output label as $y$. And then we define a template $T(·)$ that takes $x$ as input to get $x_{prompt}=T(x)=x.\ The\ intent\ is\ to\ [MASK]$. Besides, the framework needs to be well adapted and has a low computational cost during learning novel classes with a few samples. We directly use prototypes as a mapping object for each original label. Specifically, given a PLM $M$, the hidden vector at [MASK] token position of the input sentence wrapped with a template is represented as:

We represent the input sentence as $x=[w_1, ..., w_n]$, and the corresponding output label as $y$. We introduce a template function $T(\cdot)$ that takes $x$ as input and generates $x_{prompt} = T(x) =x.\ The\ intent\ is\ to\ [MASK]$. Furthermore, our framework aims to achieve effective adaptation and low computational cost when learning novel classes with limited samples. To accomplish this, we utilize prototypes as mapping objects for each original label. Specifically, given a PLM $M$, the hidden vector at the position of the [MASK] token in the input sentence wrapped with the template is represented as:
\begin{equation}
\label{eq:get_hmask}
h_{[MASK]}=M(T(x))
\end{equation}
where $h_{[MASK]}\in\mathbb{R}^h$ is the last layer's hidden representation of the PLM $M$. Then, the hidden vector in prototype space can be computed as follow:
\begin{equation}
\label{eq:get_vmask}
v_{[MASK]}=Wh_{[MASK]}
\end{equation}
where $v_{[MASK]}\in\mathbb{R}^c$ and $W$ are the encoding representation of $h_{[MASK]}$ in prototype space and a matrix for linear transformation respectively. Then we apply the cosine similarity function to calculate the similarity between two samples in prototype space, which is expressed in the following form:
\begin{equation}
\label{eq:sim}
similarity=Sim(v_i, v_j)=\frac{v_i \cdot v_j}{\Vert{v_i}\Vert\Vert{v_j}\Vert}
\end{equation}

\subsection{Knowledge Preservation}

\begin{figure}
\centering
\includegraphics[width=8.0cm]{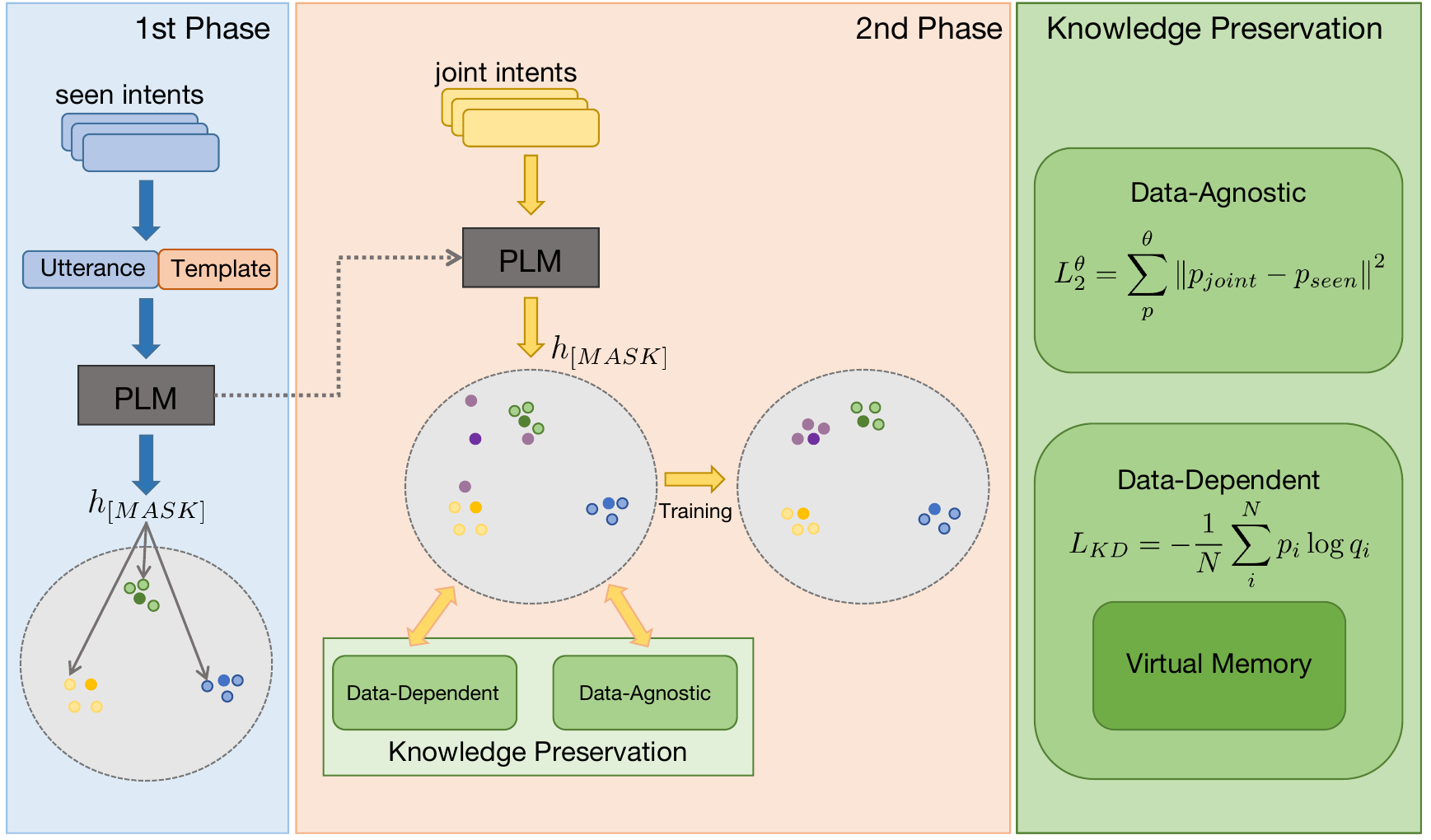}
\caption{Overview of our framework. We utilize the class incremental learning paradigm to solve the GFSID task.}
\label{fig:framework}
\end{figure}

\noindent\textbf{Data-Agnostic Knowledge Preservation.} It is critical to transfer knowledge for memorizing the prototypes of seen intents. In real-life scenarios, we cannot access past data repeatedly for some reasons such as protecting user privacy. In this case, we propose to utilize explicit weight constraints for the model, whose parameters are adaptive to seen intents. We formulate it in terms of the L2 penalization:
\begin{equation}
\label{eq:l2}
L^{\theta}_2=\sum_{p}^{\theta}{\Vert{p}_{joint}-{p}_{seen}\Vert}^2
\end{equation}
where $p_{seen}$ denotes adaptive parameters of the PLM $M$ and seen prototypes after training in the data of seen intents. $p_{joint}$ indicates the adaptive parameters excluding novel prototypes after training in the data of joint intents. The weight constraints method mentioned above forces the model to retain more old knowledge on seen intents but also allows the model to adjust the parameters to fit joint intents.

\noindent\textbf{Data-Dependent Knowledge Preservation.} In some cases, the system or the model is still permitted to access past data. Some approaches usually maintain a virtual memory for storing old data in the lifelong learning task\cite{varshney2022prompt}. And when novel classes emerge, the system will be trained with old data and novel data together. 

In this paper, we keep a virtual fixed-size memory to store the data of seen intents, whose data are randomly selected from the original dataset in a constant ratio. And to transfer the knowledge of seen intents we use knowledge distillation \cite{hinton2015distilling}. To be specific, we use the model whose parameters are adapted in seen intents to get soft labels of data stored in the memory instead of hard labels. Then when training with the data of joint intents, we distill the knowledge from the old model by minimizing the loss function in Eq. \ref{eq:kd}:

\begin{equation}
\begin{split}
L_{KD}&=-\frac{1}{N}\sum_i^N p_i\log{q_i}
% \\
% &=-\frac{1}{N}\sum_i^N\frac{\exp(v_i/\tau)}{\sum_k^N\exp(v_k/\tau)}
% \log{\frac{\exp(z_i/\tau)}{\sum_k^N\exp(z_k/\tau)}}
\end{split}
\label{eq:kd}
\end{equation}
where soft labels calculated by the old model and predicted probabilities of N classes calculated by the new model are represented as $p$ and $q$ respectively. N denotes the number of seen intents.

\subsection{Supervised Contrastive Learning}
To better understand the user's intent and achieve a superior representation, inspired by \cite{gunel2020supervised, DBLP:conf/emnlp/LiMZLLHLLC022} we utilize the supervised contrastive learning method to make a sample have high similarity with other samples of the same intent and high similarity with its prototype simultaneously. Specifically, we treat two samples belonging to the same intent as a positive pair and two samples from different intents as a negative pair. Therefore, the loss function to be minimized is as follows:
\begin{equation}
    L_{ii}=\frac{-1}{T^2}\sum_{i=1}^T\sum_{j=1}^T \log\frac{1_{y_i=y_j}\cdot\exp Sim(v_i, v_j)}{\sum_{k=1}^T\exp Sim(v_i, v_k)}
\end{equation}
where $T$ denotes the number of total instances. $v_i$ and $v_j$ are samples from the same intent. In addition, the loss function between samples and prototypes is represented as:
\begin{equation}
    L_{is}=\frac{-1}{CT}\sum_{i=1}^C\sum_{j=1}^T\log\frac{\exp Sim(v_j, c_i)}{\sum_{k=1}^C Sim(v_j, c_k)}
\end{equation}
The i-th prototype and the number of prototypes are denoted as 
$c_i$ and $C$ respectively.

\subsection{Overall Framework}
As shown in Figure \ref{fig:framework}, the entire framework is divided into two phases. In the 1st Phase, the framework learns the knowledge of seen intents with a large amount of labeled data. During the 2nd Phase, we initialize prototypes of novel intents in the prototype space and then fine-tune the parameters to fit all intents using a few instances while transferring old knowledge to new vector space with $L_2^\theta$ or Knowledge Distillation.

\noindent\textbf{1st Phase - Seen Intents Training.} We denote $\theta_{seen}$ as the parameters of PLM and seen prototypes. In order to explicitly learn the classification of seen intents, we train $\theta_{seen}$ on a great deal of labeled data of seen intents. The overall loss for the first phase is defined as:
\begin{equation}
\label{first_stage_loss}
L = L_{cls} + L_{ii} + L{is}
\end{equation}

\noindent\textbf{2nd Phase - Joint Intents Training and Knowledge Transfer.} We denote $\theta_{joint}$ as the parameters excluding novel intents prototypes during the second phase. In this phase, a few instances of seen and novel intents will emerge together. We optimize all parameters with a few instances of seen and novel intents. In the meantime, to memorize old knowledge and the prototypes of seen intents, we use different methods for different application scenarios. For scenarios where accessing old data is not allowed due to privacy policy or other issues, the optimization goal is:
\begin{equation}
\label{second_stage_l2}
L = L_{cls} + L_{ii} + L{is} + \lambda L_2^\theta
\end{equation}

For scenarios that allow access to old data, we design a virtual memory for storing a small amount of old data and then use knowledge distillation to consolidate old knowledge. The overall loss is thus:
\begin{equation}
\label{second_stage_KD}
L = L_{cls} + L_{ii} + L{is} + L_{KD}
\end{equation}

% As mentioned above, our training process consists of two stages. In each stage, we randomly initialize prototypes for the all seen intents. As shown in Eq.\ref{first_stage_loss}, during the first stage, we train these randomly initialized prototypes of seen intents and all parameters of the PLM using a large amount of labeled data. Furthermore, we further train these prototypes using two supervised contrastive learning loss functions. In the second stage, our framework employs different knowledge preservation methods based on different application scenarios (Eq.\ref{eq:kd} corresponds to scenarios where access to old data is allowed, while Eq.\ref{eq:l2} corresponds to scenarios where privacy policies restrict access to old data). Subsequently, the framework trains all parameters of PLM and prototypes of seen intents and novel intents based on Eq.\ref{second_stage_l2} and \ref{second_stage_KD}. Similarly, in the second stage, we further optimize the representation of the prototypes in the vector space using two supervised contrastive learning loss functions.

\begin{table*}
\centering
\caption{\label{main experiment results}
Experiment results of GFSID (1-shot and 5-shot) on intent detection datasets (SNIPS and NLUE).
}
\begin{tabular}{ccccccccc}
\hline
  \multirow{3}*{Model} & \multicolumn{4}{c}{\textbf{SNIPS}} & \multicolumn{4}{c}{\textbf{NLUE}} \\
 & \multicolumn{2}{c}{\textbf{1-shot}} & \multicolumn{2}{c}{\textbf{5-shot}} & \multicolumn{2}{c}{\textbf{1-shot}} & \multicolumn{2}{c}{\textbf{5-shot}} \\
& noneps & eps & noneps & eps & noneps & eps & noneps & eps \\
\hline
MN (2016) & 73.50 & 82.67 & 77.31 & 84.60 & 62.30 & 76.21 & 56.27 & 78.85\\
PN (2017) & 71.61 & 87.04 & 85.31 & 91.05 & 62.63 & 80.78 & 66.20 & 85.13\\
RN (2018) & 74.94 & 85.63 & 64.09 & 79.25 & 56.75 & 73.57 & 46.50 & 75.23\\
HATT (2019) & 71.54 & 84.51 & 86.53 & 91.85 & 64.01 & 81.39 & 67.86 & 78.41\\
MLMAN (2019) & 78.61 & 87.77 & 79.58 & 89.27 & 63.12 & 82.65 & 60.70 & 84.45\\
HAPN (2019) & 74.33 & 85.37 & 86.19 & 89.40 & 60.44 & 82.00 & 68.34 & 84.75\\
SMAN (2020) & 81.85 & 88.10 & 87.87 & 93.18 & 66.10 & 89.54 & 72.18 & 87.76\\
MLADA (2021) & 71.15 & 79.14 & 88.90 & 92.89 & 45.31 & 67.91 & 61.93 & 82.56\\
DFEPN (2022) & 92.30 & \textbf{95.41} & 93.42 & 96.27 & 72.09 & 87.35 & 82.28 & 93.17\\
\hline 
\textbf{DDKP} (ours) & \textbf{94.11} & 93.99 & 96.33 & \textbf{96.91} & \textbf{82.27} & \textbf{89.91} & \textbf{88.77} & \textbf{93.94}\\
\textbf{DAKP} (ours) & 94.10 & 94.18 & \textbf{96.58} & 96.85 & 80.75 & 89.75 & 88.29 & 93.66 \\
\hline
\end{tabular}
\end{table*}

\section{EXPERIMENT}
\label{sec:typestyle}

Following the previous methodology \cite{nguyen2020dynamic} we perform extensive experiments on SNIPS and NLUE datasets to demonstrate the effectiveness of our framework. We conduct performance evaluations of GFSID on two datasets. The GFSID testing samples comprise 20\% instances from the seen intents and all instances from the novel intents. The remaining 80\% of samples from seen intents are used for training. For testing, support instances are pre-selected for both the 1-shot and 5-shot settings. Besides, we compare the proposed approach with strong baselines\cite{gao2019hybrid, sun2019hierarchical, ye2019multi, snell2017prototypical, vinyals2016matching, sung2018learning, nguyen2020dynamic, ijcai2022p617, han2021meta}. All performance results of the methods selected for comparison are taken from \cite{ijcai2022p617}.

Our framework is implemented using PyTorch and the OpenPrompt toolkit \cite{ding2022openprompt}. $RoBERTa_{base}$ \cite{liu2019roberta} is used as the text encoder. In the first Phase, we utilize the Adam optimizer with a learning rate of $1e-5$ and train with a batch size of 64. In the second Phase, the learning rate is $1e-4$. For SNIPS, the batch size is set to 5. Training all parameters for 20 epochs. For NLUE, we set the batch size to 16 and train all parameters for 15 epochs. Following previous approach \cite{nguyen2020dynamic}, we evaluate all methods in two different settings: Non-episodic Evaluation (noneps) and Episodic Evaluation (eps).

\textbf{Episodic Evaluation:} 1000 episodes for C-way K-shot testing, where C is 4 or 10 for SNIPS and NLUE, respectively. Query set consists of 5 samples for each intent.

\textbf{Non-episodic Evaluation:} Conduct a single round of testing on all unlabeled samples, which is more challenging and realistic.

\subsection{Experimental Results}

Our primary findings are presented in Table \ref{main experiment results}. For the SNIPS dataset, we computed the average accuracy across 3 distinct random seeds for 3 experimental groups. To evaluate our framework on the NLUE dataset, we employed 10-fold cross-validation. The experimental results demonstrate that our approach consistently outperforms previous methods in the majority of scenarios. Particularly in the more challenging and realistic noneps setting, our method demonstrates significant improvements of 10.18\% and 6.49\% for 1-shot and 5-shot respectively, as compared to the baseline. In the eps setting, our method achieve improvements of 2.56\% and 0.77\% in the 1-shot and 5-shot setting respectively, on the NLUE dataset. On the SNIPS dataset, our method exhibit a 0.64\% improvement in the 5-shot setting. Additionally, employing the data-dependent knowledge preservation method yields superior performance in most cases.

\subsection{Ablation Study}
\begin{table}
\caption{\label{knowledge preservation}
The comparison of two knowledge preservation methods and directly learning without any knowledge preservation. Results on SNIPS dataset (1-shot and 5-shot noneps).
}
\centering
\begin{tabular}{ccc}
\hline
Method & 1-shot & 5-shot \\
\hline
DDKP & 94.11 & 96.33 \\
DAKP & 94.10 & 96.58 \\
w/o $L_{KD}$ w/o $L_{2}^\theta$ & 88.98 & 95.49 \\
\hline
\end{tabular}
\end{table}

\begin{figure}
\centering
\includegraphics[width=7.8cm]{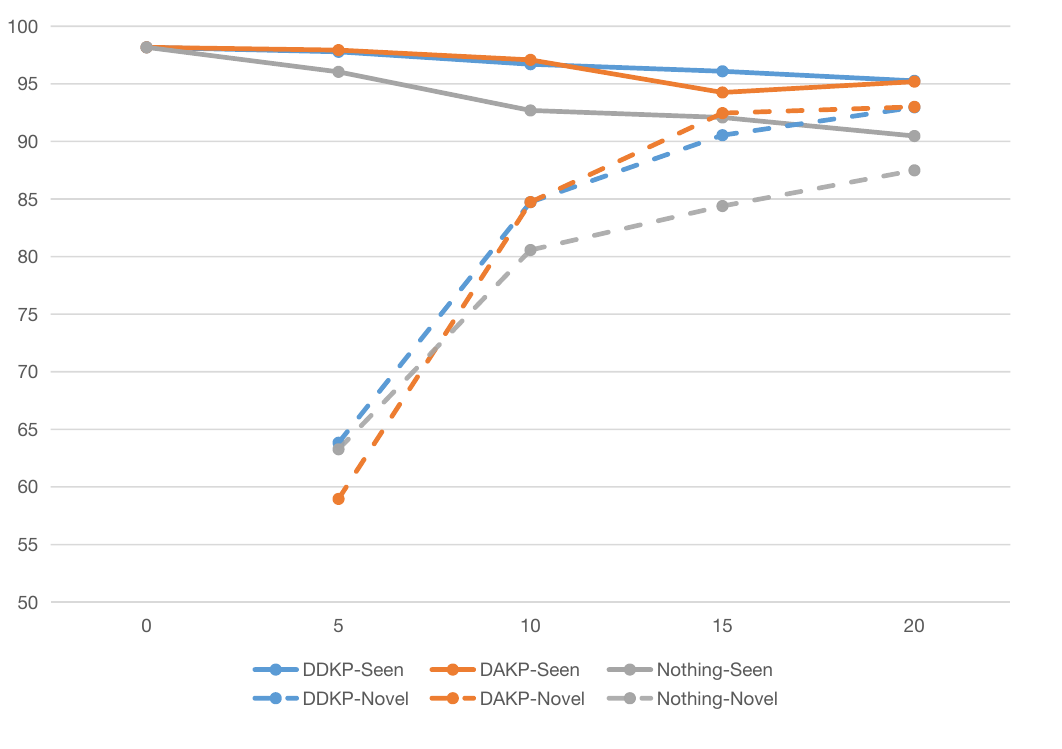}
\caption{Investigating old knowledge preservation performance of DDKP and DAKP when learning novel intents. }
\label{fig:ablation}
\end{figure}

The observations from Figure \ref{fig:ablation} and Table \ref{knowledge preservation} are as follows:

(1) Our proposed DDKP and DAKP play a crucial role in preserving old knowledge, and the effects are significant.

(2) DDKP provides more stable retention of old knowledge. This is because DAKP directly relies on parameter constraints to achieve knowledge preservation, which leads to some performance loss. 

(3) DDKP and DAKP have an advantage when learning novel intents knowledge. This is because seen intents and novel intents share common knowledge, and the retained common knowledge can assist in learning novel intents knowledge.

\section{CONCLUSION}
\label{sec:majhead}

In this paper, we innovatively introduce a conversion of the GFSID task into the class incremental learning paradigm to extend it to a generalized setup. We propose a straightforward and efficient two-stage learning framework for sequentially acquiring knowledge of various intents. Additionally, to facilitate the transfer of intent knowledge across different stages, we introduce two knowledge preservation methods that closely align with realistic applications. We conduct extensive experiments and demonstrate the effectiveness of our proposed framework through results obtained from two real-world intent detection datasets.

% \section{REFERENCES}
% \label{sec:refs}

% List and number all bibliographical references at the end of the
% paper. The references can be numbered in alphabetic order or in
% order of appearance in the document. When referring to them in
% the text, type the corresponding reference number in square
% brackets as shown at the end of this sentence \cite{C2}. An
% additional final page (the fifth page, in most cases) is
% allowed, but must contain only references to the prior
% literature.

\newpage
\bibliographystyle{IEEEbib}
\bibliography{refs}

\end{document}